\documentclass{article} 
\usepackage{iclr2026_conference,times}

\usepackage{amsmath,amsfonts,bm}

\newcommand{\newterm}[1]{{\bf #1}}

\def\Figref#1{Figure~\ref{#1}}

\def\Secref#1{Section~\ref{#1}}

\def\eqref#1{equation~\ref{#1}}

\def\1{\bm{1}}

\def\vtheta{{\bm{\theta}}}

\def\vg{{\bm{g}}}

\DeclareMathAlphabet{\mathsfit}{\encodingdefault}{\sfdefault}{m}{sl}
\SetMathAlphabet{\mathsfit}{bold}{\encodingdefault}{\sfdefault}{bx}{n}

\def\gB{{\mathcal{B}}}

\def\gD{{\mathcal{D}}}
\def\gE{{\mathcal{E}}}

\def\gJ{{\mathcal{J}}}

\newcommand{\E}{\mathbb{E}}

\usepackage[utf8]{inputenc} 
\usepackage[T1]{fontenc}    
\usepackage[table]{xcolor}
\definecolor{linknavy}{RGB}{35,70,115}
\usepackage{hyperref}       
\hypersetup{
    colorlinks=true,
    linkcolor=linknavy,
    citecolor=linknavy,
    urlcolor=linknavy
}
\usepackage{url}            
\usepackage{booktabs}       
\usepackage{amsfonts}       
\usepackage{nicefrac}       
\usepackage{microtype}      
\usepackage{colortbl}

\usepackage{multirow}
\usepackage{graphicx}
\usepackage{amsmath}
\usepackage{amssymb}
\usepackage{amsthm}
\usepackage{bm}

\newtheorem{theorem}{Theorem}
\newtheorem{proposition}{Proposition}

\newtheorem{definition}{Definition}

\usepackage{subcaption}
\usepackage{enumitem}

\usepackage{algorithm}
\usepackage{algpseudocode}

\usepackage{wrapfig}

\newcommand{\method}{\mbox{SARA}}              
\newcommand{\peff}{p^{\mathrm{eff}}}           
\newcommand{\Bud}{N}                           
\newcommand{\taulow}{\tau_{\mathrm{low}}}      
\newcommand{\ninit}{n_{0}}                     
\newcommand{\Betaf}{\mathrm{B}}                
\newcommand{\Beta}{\mathrm{Beta}}              
\newcommand{\Bin}{\mathrm{Bin}}                
\newcommand{\Bernoulli}{\mathrm{Bernoulli}}

\definecolor{saracol}{RGB}{197,90,17}

\title{Early Verdicts, Better Budgets: Sequential Adaptive Rollout Allocation for Compute-Efficient RLVR}

\iclrfinalcopy

\author{
  Pixel Nomand$^{1}$\quad
  Elena Voss$^{1}$\quad
  Marcus Hale$^{2}$\quad
  Sofia Reyes$^{1}$\\
  $^{1}$University of Wisconsin--Madison\\
  $^{2}$University of Washington
}

\newcommand{\saraRollSave}{22\%}
\newcommand{\saraTokSave}{23\%}
\newcommand{\saraCompRollSave}{67\%}

\newcommand{\saraEffFrac}{26\%}
\newcommand{\saraLostRate}{8.6\%}
\newcommand{\saraGroupK}{8}
\newcommand{\saraProbe}{2}
\newcommand{\saraTau}{0.45}
\newcommand{\saraSteps}{300}
\newcommand{\saraBatchB}{64}

\begin{document}

\maketitle

\begin{abstract}
Reinforcement learning with verifiable rewards (RLVR) is bottlenecked by rollout
generation, yet many sampled prompts produce \emph{saturated} groups (all
responses correct or all incorrect) whose zero reward variance yields no
policy-gradient signal. Existing remedies either oversample a larger candidate
pool and discard saturated prompts (dynamic sampling), paying heavy extra
rollouts, or predict prompt difficulty \emph{before} sampling, which is fragile
under a shifting policy. We observe that a group's effectiveness is usually
\emph{decided early}, within the first few of its rollouts, so spending a full
group on an already-decided prompt is wasteful. We cast per-step rollout
collection as a budget-constrained \emph{sequential allocation} (optimal
stopping) problem and introduce \method{} (Sequential Adaptive Rollout
Allocation). \method{} maintains a Beta posterior over each prompt's success
rate, evaluates a closed-form predictor of group effectiveness, and applies a
two-threshold, SPRT-style rule that commits effective groups, abandons saturated
ones after a short probe, and reallocates the freed budget to fresh prompts,
without any extra prediction rollouts. We prove abandonment reliability,
expected rollout savings, fixed-budget yield dominance, and a link between
effective-group yield and the GRPO gradient norm. On mathematical reasoning and
planning with 1.5B/3B models on a single GPU, \method{} matches DPS (both below
the DS oracle) while using \saraRollSave{} fewer rollouts than DS; composing
\method{} with DPS yields the best accuracy, slightly above DS, at
\saraCompRollSave{} fewer rollouts (near-uniform cost).
\end{abstract}

\section{Introduction}
\label{sec:intro}

Reinforcement learning with verifiable rewards (RLVR) has become the central
post-training mechanism for eliciting reasoning in large language models (LLMs),
from mathematical problem solving to multi-step planning and
code~\citep{jaech2024o1,guo2025deepseekr1,kimi2025k15,shao2024deepseekmath,
lightman2023verify}. Its dominant cost is \emph{rollout generation}: each policy
update consumes many long chain-of-thought samples~\citep{chen2025longcot,
sui2025overthinking}, and deciding where to spend a limited rollout budget is a
central design problem~\citep{zheng2025selective,lin2025cppo,aggarwal2025l1}.

\begin{figure}[t]
\centering
\includegraphics[width=\textwidth]{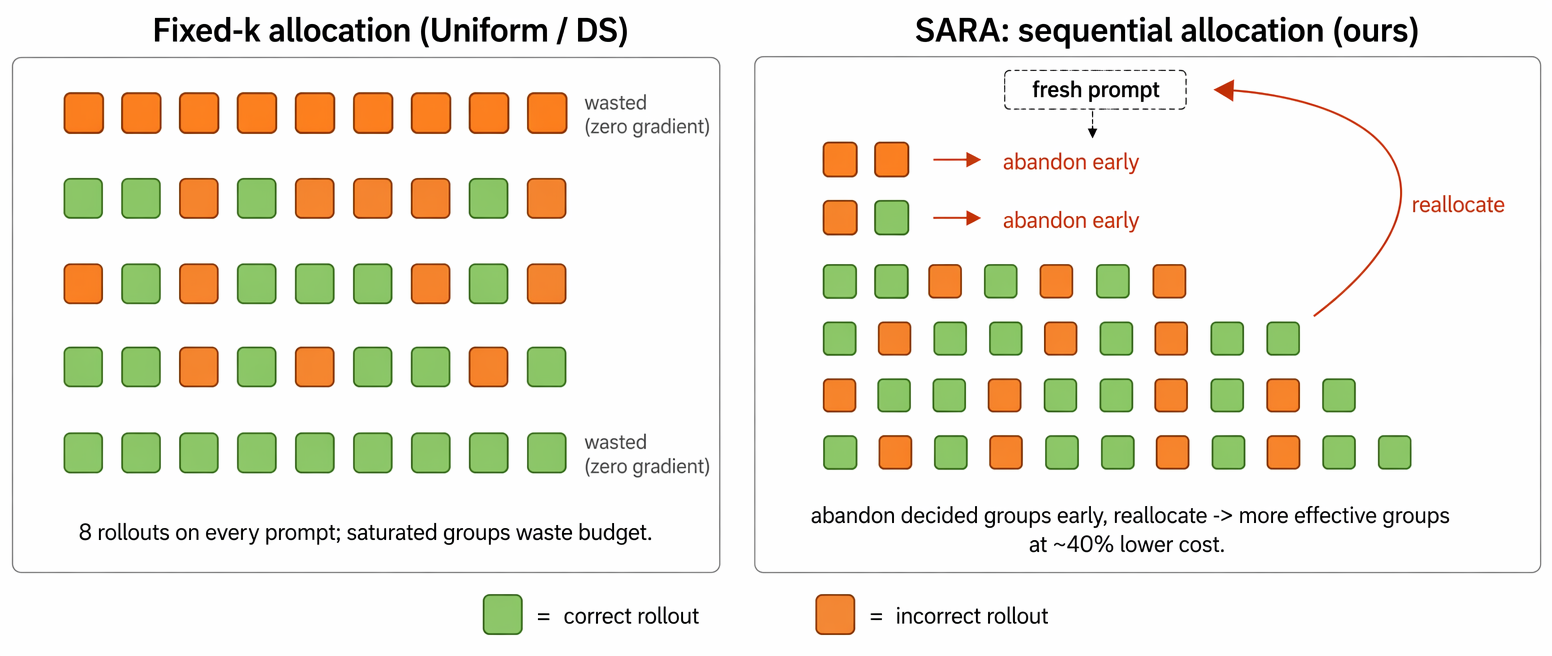}
\caption{\textbf{Sequential allocation spends rollouts where verdicts are
contrastive.} \emph{Left:} fixed-$k$ allocation (uniform sampling, or dynamic
sampling after filtering) generates a full group of $k$ rollouts for every
prompt; \emph{saturated} groups (all-correct or all-incorrect, faded) have zero
reward variance and contribute no GRPO gradient. \emph{Right:} \method{}
abandons a group as soon as its outcome is statistically decided and reallocates
the freed budget to fresh prompts, assembling more effective (mixed) groups at
$\sim$\saraRollSave{} lower cost.}
\label{fig:teaser}
\end{figure}

A recurring obstacle is that prompts contribute very unevenly. Under group-based
estimators such as GRPO~\citep{shao2024deepseekmath} and related
variants~\citep{hu2025reinforcepp,liu2025drgrpo,liu2025prorl}, a prompt whose
sampled responses are \emph{all} correct or \emph{all} incorrect produces a
zero-variance reward group and hence a vanishing normalized
advantage~\citep{yu2025dapo,bae2025online}. Such \emph{saturated} groups are far
from rare: under uniform sampling they routinely make up a majority of a batch
(\Figref{fig:motivation}a), so much of the rollout budget is spent producing
gradients that are exactly zero. Only \emph{effective} groups (those with a
mixed outcome) actually drive learning.

Two families of methods attack this waste. Evaluate-then-filter methods such as
dynamic sampling (DS) in DAPO~\citep{yu2025dapo} and online difficulty
filtering~\citep{bae2025online} oversample a larger candidate pool, generate full
groups for all of them, and discard the saturated ones; this guarantees a clean
batch but multiplies rollout cost and often dominates
training~\citep{zheng2025selective,lin2025cppo}. Offline curation instead
pre-selects informative prompts by static difficulty or
diversity~\citep{li2025limr,ye2025limo,wang2025oneshot}, but cannot track a
shifting online policy. Predict-then-select methods estimate each prompt's
difficulty \emph{before} sampling~\citep{chen2025sec,mao2026dps,
qu2026gps} and prioritize the promising ones; this adds no extra rollouts but
stakes the batch on a forecast, so when the policy shifts quickly or histories
are sparse the predictions err and the batch is polluted
again~\citep{zhang2025srpo,zheng2025selective}. The two families sit at opposite
ends of a \emph{prediction-versus-verification} trade-off: predict a prompt's
label before paying for it, or verify it by paying in full. Both decide at the
prompt level, before the group's own rollouts can inform the choice (as do
selective generation and tree-structured
allocators~\citep{zheng2025selective,zou2026trace}), and so neither exploits the
cheap evidence sitting inside the group itself.

That evidence is the starting point for this work. Within a group, the
effectiveness verdict is typically reached long before the group is complete:
once a prompt yields one correct \emph{and} one incorrect response it is
effective forever, since a mixed group cannot become unmixed, while a run of
identical outcomes quickly concentrates the posterior over the success rate near
$0$ or $1$. Empirically (\Figref{fig:mechanism}a), the eventual label of most
groups is settled after only a handful of their $k$ rollouts, so spending the
full $k$ on an already-decided prompt wastes compute. Reaching this verdict needs
no extra model calls; it simply reads the rollouts the optimizer would generate
anyway. The same early-stopping intuition underlies classical sequential
analysis~\citep{wald1945sequential,robbins1952sequential}, which we adapt to
per-step RLVR budgeting.

We therefore recast a single RLVR step as a budget-constrained \emph{sequential
allocation} problem over a stream of prompts, and instantiate it as \method{}
(Sequential Adaptive Rollout Allocation). \method{} probes prompts in batched
rounds; after each round it updates a Beta posterior over every active prompt's
success rate, evaluates a closed-form posterior-predictive probability that the
group will be effective at full size, and applies a two-threshold,
sequential-probability-ratio-style rule: \emph{commit} a group once it is mixed,
\emph{abandon} it once it is almost surely saturated, and otherwise
\emph{continue}. Abandoning a likely-saturated prompt after a short probe frees
rollouts that are reallocated to fresh prompts, so a fixed budget assembles more
effective groups (\Figref{fig:teaser}). \method{} is orthogonal to prompt
selection and composes with it, adds no prediction rollouts, and introduces only
a handful of synchronization rounds.

\paragraph{Contributions.}
(1) We identify \emph{early decidability} of group effectiveness and reframe
per-step rollout collection as sequential allocation / optimal stopping, a
budget axis orthogonal to prompt-level selection. (2) We derive a closed-form
Beta--Binomial effectiveness predictor and a two-threshold stopping rule,
yielding \method{}, a prediction-rollout-free allocator that drops into any
GRPO-style pipeline. (3) We prove abandonment reliability, expected rollout
savings, fixed-budget yield dominance over uniform allocation, and a lower bound
relating effective-group yield to the expected squared GRPO gradient. (4) On
1.5B/3B models with a single GPU, \method{} matches DPS while using
\saraRollSave{} fewer rollouts than DS, and \method{}$+$DPS edges DS at
\saraCompRollSave{} fewer rollouts.

\section{Preliminaries and Problem Setup}
\label{sec:prelim}

\paragraph{RLVR and GRPO.}
Let $q\sim\gD$ be a prompt drawn from a dataset and $o\sim\pi_\vtheta(\cdot\mid
q)$ a response from policy $\pi_\vtheta$. RLVR maximizes the expected verifiable
reward $\max_{\vtheta}\,\E_{q\sim\gD,\,o\sim\pi_\vtheta(\cdot\mid q)}\,[r(q,o)]$,
where $r(q,o)\in\{0,1\}$ checks correctness~\citep{guo2025deepseekr1,
lightman2023verify}. Relative to PPO~\citep{schulman2017ppo}, Group Relative
Policy Optimization (GRPO)~\citep{shao2024deepseekmath} avoids a value network by
sampling, for each prompt $q$, a group of $k$ responses $\{o_i\}_{i=1}^{k}$ and
normalizing rewards \emph{within} the group:
\begin{equation}
\hat{A}_i \;=\; \frac{r(q,o_i)-\mathrm{mean}(\{r(q,o_j)\}_{j=1}^{k})}
{\mathrm{std}(\{r(q,o_j)\}_{j=1}^{k})},\qquad
\gJ(\vtheta)=\E\Big[\textstyle\frac1k\sum_i \min\big(\rho_i\hat A_i,\,
\mathrm{clip}(\rho_i,1{\pm}\epsilon)\hat A_i\big)\Big],
\label{eq:grpo}
\end{equation}
with importance ratio $\rho_i=\pi_\vtheta(o_i\mid q)/\pi_{\vtheta_{\mathrm{old}}}(o_i\mid q)$.
Subsequent variants retain this within-group normalization while adjusting
stability, length bias, or the critic-free
estimator~\citep{yu2025dapo,hu2025reinforcepp,liu2025drgrpo,liu2025prorl}.

\begin{figure}[t]
\centering
\includegraphics[width=\textwidth]{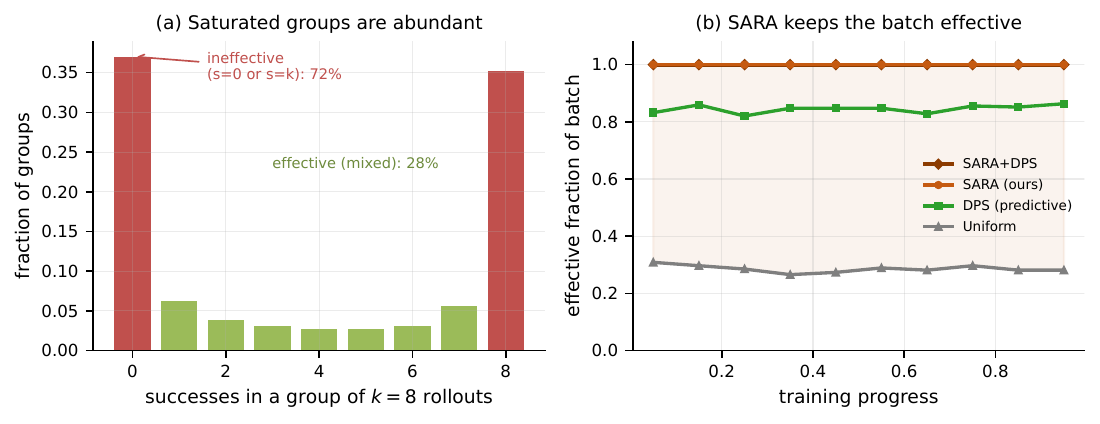}
\caption{\textbf{Saturated groups dominate, and \method{} keeps the batch
effective.} \emph{(a)} Distribution of per-group success counts under uniform
sampling ($k{=}8$): most groups sit at $s{=}0$ or $s{=}k$ (zero gradient).
\emph{(b)} Fraction of the \emph{training batch} that is effective over
training: \method{} (and the dynamic-sampling oracle) deliver near-fully
effective batches; predictive selection recovers most of it but degrades as the
policy shifts; uniform sampling leaves the majority of the batch wasted.}
\label{fig:motivation}
\end{figure}

\paragraph{Effective and saturated groups.}
For a prompt $q$ let $\gamma_q=\E[r(q,o)]\in[0,1]$ be its (latent, policy- and
step-dependent) \newterm{success rate}. A group of $k$ rollouts has success
count $S_k\sim\Bin(k,\gamma_q)$.
\begin{definition}[Effective group]
\label{def:effective}
A group is \newterm{effective} if its success count is mixed, $1\le S_k\le k-1$,
and \newterm{saturated} otherwise ($S_k\in\{0,k\}$).
\end{definition}
A saturated group has zero reward variance, so every $\hat A_i$ in
Eq.~\eqref{eq:grpo} is $0$ and the group contributes \emph{no} gradient. The
probability that a group is effective is
$\Phi(\gamma_q)\,{=}\,1-\gamma_q^{k}-(1-\gamma_q)^{k}$, which is near $0$ for the
many easy ($\gamma_q\!\to\!1$) and hard ($\gamma_q\!\to\!0$) prompts and peaks
at intermediate difficulty. Consequently, under uniform sampling a large
fraction of each batch is saturated and wasted (\Figref{fig:motivation}a;
cf.~\citealp{yu2025dapo,bae2025online,zheng2025selective}).

\paragraph{Per-step budget and the two existing remedies.}
A vanilla GRPO step draws $B$ prompts and generates $k$ rollouts each, spending a
budget of $\Bud=B\cdot k$ rollouts but obtaining only $\Phi\cdot B$ effective
groups on average. \emph{Dynamic sampling} (DS/DAPO)~\citep{yu2025dapo} and
related online filters~\citep{bae2025online,zhang2025srpo} over-sample a
candidate pool $\hat\gB$ (with $|\hat\gB|\!>\!B$), generate full groups for all
candidates, and keep the effective ones,
$\gB_t=\{q\in\hat\gB:\mathrm{std}(\{r(q,o_i)\})>0\}$; this yields a clean batch
of $B$ effective groups but at expected cost $\approx Bk/\Phi$ rollouts, several
times $\Bud$. \emph{Predictive selection}~\citep{chen2025sec,
mao2026dps,qu2026gps} estimates $\gamma_q$ from history and samples prompts with
intermediate predicted difficulty, spending no extra rollouts but risking a
polluted batch when predictions err. Length-control methods instead shrink
tokens \emph{within} each
rollout~\citep{aggarwal2025l1,hou2025thinkprune,fatemi2025concise}; they are
complementary to deciding \emph{how many} rollouts to collect. We target a clean
batch of $B$ effective groups while spending far fewer rollouts than DS and
remaining robust to the prediction errors that limit predictive selection.

\paragraph{Goal.}
At each step, assemble $B$ effective groups for the GRPO update, minimizing the
rollouts (and tokens) spent, using \emph{only} the outcomes of the rollouts
themselves (no auxiliary evaluation passes).

\section{Sequential Adaptive Rollout Allocation}
\label{sec:method}

\subsection{Rollout collection as sequential allocation}
\label{sec:method:formulation}
Within one RLVR step we receive a stream of prompts and must, under a rollout
budget, assemble $B$ effective groups. The $k$ rollouts of a prompt need not be
generated at once: we may generate a few, inspect their outcomes, and decide
whether to invest more in this prompt or move on. This is a finite-horizon
sequential decision problem. For prompt $q$ with $n$ observed rollouts and $s$
successes so far, we choose among \{commit, abandon, continue\}; abandoning frees
budget for other prompts. The objective is to maximize the number of effective
groups obtained per rollout (equivalently, to minimize the rollouts spent to
obtain $B$ effective groups), using only the observed outcomes $(n,s)$. We show
this admits a near-optimal index policy (App.~\ref{app:proofs}); \method{} is its
tractable, closed-form realization.

\paragraph{Prior methods as boundary cases.}
Viewed this way, existing efficiency methods are degenerate \emph{stopping
policies} on the same decision tree. Uniform sampling and dynamic sampling never
stop early: they commit every prompt to a full group of $k$ rollouts and only
then read its label, so they pay for saturated groups in full. Predict-then-select
methods stop at $n{=}0$, before any rollout, by acting on a difficulty forecast,
trading rollout cost for prediction risk. These sit at opposite ends of one axis:
\emph{how much evidence to gather before deciding}. The Bayes-optimal amount of
evidence is neither ``none'' (blind prediction) nor ``all $k$'' (no early
stopping), but a data-dependent stopping time. \method{} gathers just enough
evidence to decide each group and no more, which turns filter vs.\ predict vs.\
allocate into one sequential-decision problem.

\subsection{Early decidability of group effectiveness}
\label{sec:method:predict}
We place a Beta prior $\gamma_q\sim\Beta(\alpha_0,\beta_0)$ on each success rate
(uniform $\alpha_0{=}\beta_0{=}1$ by default; an informative prior from history
is an option, \Secref{sec:exp}). After observing $n$ rollouts with $s$
successes, conjugacy gives the posterior
$\gamma_q\mid(n,s)\sim\Beta(\alpha_0{+}s,\beta_0{+}n{-}s)$. The quantity that
governs the decision is the posterior-predictive probability that the
\emph{completed} group of size $k$ will be effective.

\begin{proposition}[Closed-form effectiveness predictor]
\label{prop:predictor}
Let $r=k-n$ remaining rollouts. The posterior-predictive probability of an
effective group is
\begin{equation}
\peff(n,s)\;=\;
\begin{cases}
1, & s\ge 1 \text{ and } n-s\ge 1 \ \text{(already mixed)},\\[2pt]
1-\dfrac{\Betaf(\alpha_0,\,\beta_0+n+r)}{\Betaf(\alpha_0,\,\beta_0+n)}, & s=0,\\[8pt]
1-\dfrac{\Betaf(\alpha_0+n+r,\,\beta_0)}{\Betaf(\alpha_0+n,\,\beta_0)}, & s=n,
\end{cases}
\label{eq:peff}
\end{equation}
where $\Betaf(\cdot,\cdot)$ is the Beta function. With a uniform prior and
$s{=}0$ this simplifies to $\peff(n,0)=1-\frac{n+1}{k+1}=\frac{k-n}{k+1}$
(symmetrically for $s{=}n$).
\end{proposition}

\begin{figure}[t]
\centering
\includegraphics[width=\textwidth]{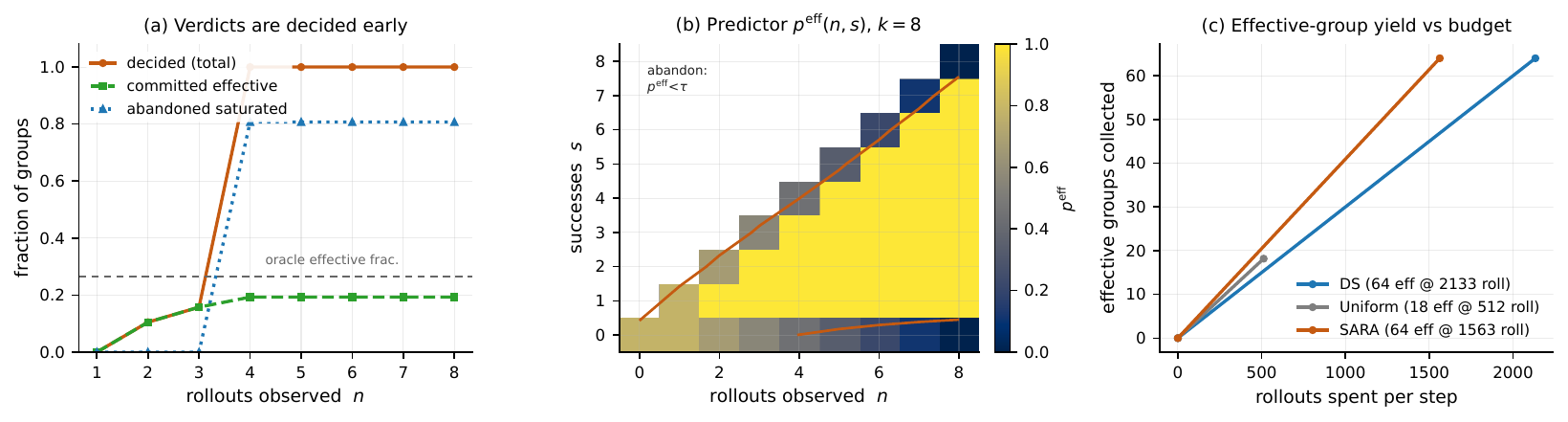}
\caption{\textbf{Mechanism of \method{}.} \emph{(a)} Fraction of groups whose
final effectiveness label is already decided after $n$ observed rollouts
(committed-effective $+$ confidently-abandoned); most verdicts are reached well
before $n{=}k$, and they agree with the oracle label. \emph{(b)} The closed-form
predictor $\peff(n,s)$ of Eq.~\eqref{eq:peff} with the abandon region
($\peff<\taulow$) outlined. \emph{(c)} Effective groups collected versus
rollouts spent: \method{} reaches a full clean batch at a fraction of the cost
of dynamic sampling.}
\label{fig:mechanism}
\end{figure}

\begin{proposition}[Irreversibility and monotone decidability]
\label{prop:monotone}
Effectiveness is absorbing: once a group is mixed it stays effective regardless
of further rollouts. Along an all-fail (resp.\ all-pass) prefix, $\peff(n,s)$ is
strictly decreasing in $n$. Hence each prompt is \emph{decided} at a
well-defined stopping time, and the fraction of groups decided after $n$
rollouts increases monotonically to $1$ (\Figref{fig:mechanism}a).
\end{proposition}

\paragraph{Two-threshold sequential rule.}
Given $\peff(n,s)$, \method{} applies a sequential test reminiscent of Wald's
SPRT~\citep{wald1945sequential}, with a single lower threshold $\taulow$ because
the upper decision (mixed $\Rightarrow$ effective) is exact:
\begin{equation}
\textsc{decide}(n,s)=
\begin{cases}
\textsc{commit} & \text{if the group is mixed (and size $\ge$ commit floor)},\\
\textsc{abandon} & \text{if } \peff(n,s)<\taulow,\\
\textsc{continue} & \text{otherwise (sample one more rollout)}.
\end{cases}
\label{eq:decide}
\end{equation}
A committed group enters the training batch; an abandoned prompt is dropped and
its remaining budget is reallocated. The threshold $\taulow$ is the single knob
trading rollout savings against the risk of abandoning a would-be-effective
prompt (analyzed below and in \Secref{sec:exp}).

\subsection{The \method{} algorithm}
\label{sec:method:algo}
\begin{figure}[t]
\centering
\includegraphics[width=0.86\textwidth]{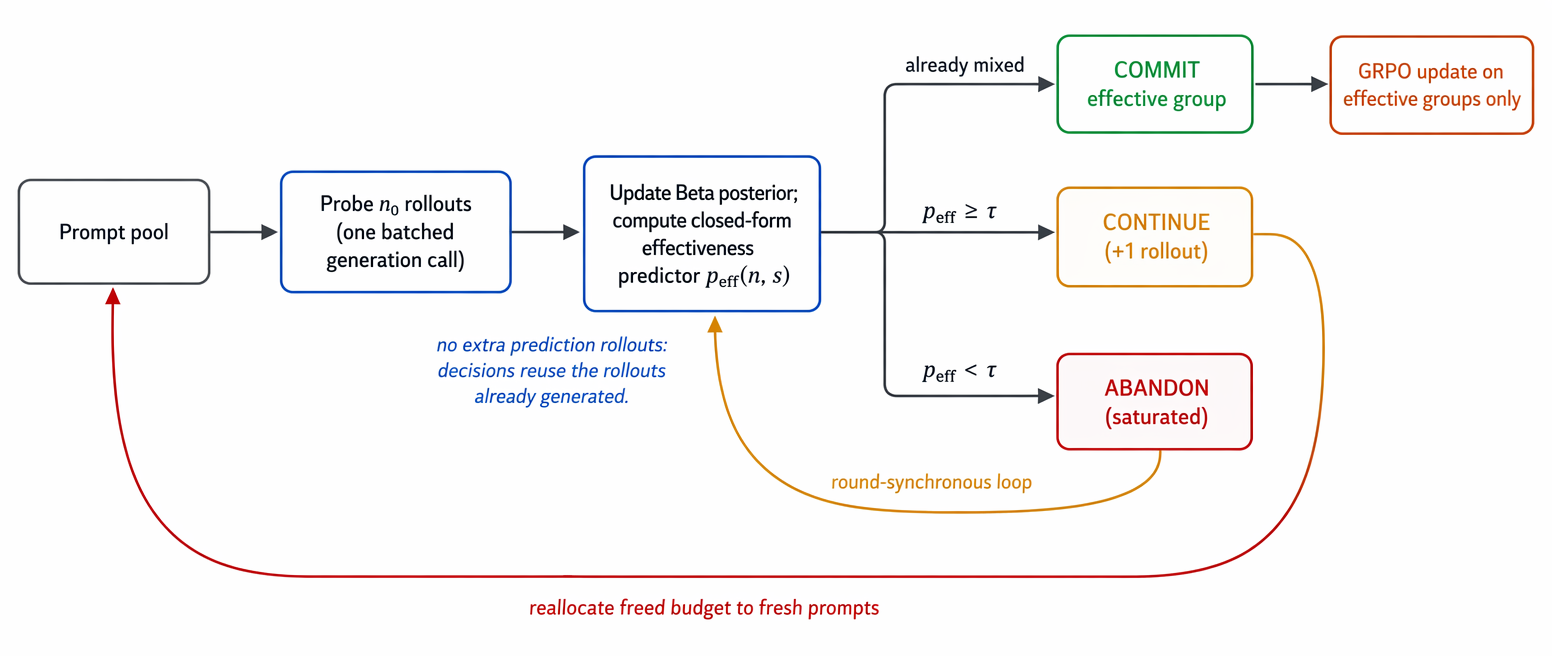}
\caption{\textbf{\method{} inside one GRPO step.} Prompts are probed in batched
rounds; a Beta posterior and the closed-form predictor $\peff$ route each group
to \textsc{commit}/\textsc{continue}/\textsc{abandon}; abandoned budget is
reallocated to fresh prompts, and only effective groups reach the update. No
auxiliary prediction rollouts are used.}
\label{fig:arch}
\end{figure}

\method{} runs in batched, round-synchronous fashion (\Figref{fig:arch}): each
round issues a \emph{single} batched generation call over all active prompts
(preserving inference throughput), then applies Eq.~\eqref{eq:decide} and
reallocates. Committed prompts leave the working set; abandoned prompts free
budget that is used to draw fresh prompts; continuing prompts, prioritized by a
yield index (we use $\peff$, a one-step look-ahead toward effective groups),
receive one more rollout. The loop ends when $B$ effective groups are collected
or the safety budget is exhausted. Algorithm~\ref{alg:sara} summarizes one step.

\begin{algorithm}[t]
\caption{\method{}: one RLVR step (round-synchronous)}
\label{alg:sara}
\begin{algorithmic}[1]
\Require prompt pool $\gD$; target $B$ effective groups; group size $k$; probe
$\ninit$; threshold $\taulow$; prior $(\alpha_0,\beta_0)$; budget cap $\Bud$.
\State $\mathcal{C}\gets\emptyset$ (collected); $\mathcal{A}\gets$ probe $B$
fresh prompts with $\ninit$ rollouts each \Comment{one batched call}
\While{$|\mathcal{C}|<B$ \textbf{and} budget left and $\mathcal{A}\neq\emptyset$}
  \For{each active prompt $q\in\mathcal{A}$ with stats $(n_q,s_q)$}
    \State $d\gets\textsc{decide}(n_q,s_q)$ via Eq.~\eqref{eq:decide}
    \If{$d=\textsc{commit}$} move $q$ to $\mathcal{C}$
    \ElsIf{$d=\textsc{abandon}$} drop $q$ \Comment{free its budget}
    \EndIf
  \EndFor
  \State \textbf{reallocate:} draw fresh prompts (probe $\ninit$) to refill the
  working set toward the remaining need
  \State advance the continuing prompts by one rollout, prioritized by
  $\peff$ \Comment{one batched call}
\EndWhile
\State \Return $\mathcal{C}$; perform a GRPO update (Eq.~\ref{eq:grpo}) on the
effective groups in $\mathcal{C}$
\end{algorithmic}
\end{algorithm}

\paragraph{Cost.}
The only overhead beyond generation is, per active prompt per round, a Beta
update and one evaluation of Eq.~\eqref{eq:peff} (a few $\log\Gamma$
evaluations): $O(1)$ time and memory, negligible against a single LLM rollout,
and requiring zero extra model calls. The number of synchronization rounds is
$O(k)$ in the worst case but small in practice (most groups are decided in
$2$--$4$ rounds, \Figref{fig:mechanism}a).

\subsection{Theoretical guarantees}
\label{sec:method:theory}
We summarize four results; full statements and proofs are in
App.~\ref{app:proofs}. The first bounds the only failure mode of early
abandonment.
\begin{theorem}[Abandonment reliability]
\label{thm:reliability}
The posterior probability that an abandoned group would have been effective is
exactly $\peff<\taulow$. Hence the expected number of effective groups lost per
step is at most $\taulow\cdot|\text{abandoned}|$, and choosing
$\taulow\to0$ makes \method{} lossless in the limit.
\end{theorem}
\begin{theorem}[Expected rollout savings]
\label{thm:savings}
For a prompt with success rate $\gamma$, the expected rollouts \method{} spends
before deciding is at most $\min\{k,\,n^\star(\gamma)\}$ with
$n^\star(\gamma)\!=\!O\!\big(\tfrac{\log(1/\taulow)}{\log(1/\max(\gamma,1-\gamma))}\big)$
for saturated prompts, strictly less than $k$. Aggregating over the difficulty
distribution, the expected cost to assemble $B$ effective groups is strictly
below the $\approx Bk/\Phi$ of dynamic sampling, and the gap grows with $k$.
\end{theorem}
\begin{theorem}[Fixed-budget yield dominance]
\label{thm:yield}
At a fixed rollout budget, the expected number of effective groups assembled by
\method{} is at least that of uniform fixed-$k$ allocation, with strict
inequality whenever the success-rate distribution is non-degenerate.
\end{theorem}
\begin{theorem}[Effective yield lower-bounds the gradient]
\label{thm:gradient}
Saturated groups contribute exactly zero to the GRPO gradient. Consequently the
expected squared gradient norm obeys
$\E\|\nabla_\vtheta\gJ\|^2\ge c\cdot\E[\,\#\text{effective groups}\,]$ for a
constant $c>0$ depending on the per-group gradient energy. Maximizing
effective-group yield thus maximizes a lower bound on the per-step update
magnitude.
\end{theorem}
Together, Thms.~\ref{thm:reliability}--\ref{thm:gradient} say that \method{} is
reliable, cheaper than oversample-and-filter, never worse than uniform at fixed
budget, and optimizes a quantity tied to learning progress.
App.~\ref{app:proofs:gittins} casts the step as a restless bandit and shows that
the two-threshold rule of Eq.~\eqref{eq:decide} is the Bayes-optimal stopping
boundary under a single-prompt relaxation and the myopic index policy in general.

\section{Experiments}
\label{sec:exp}

We evaluate whether sequential in-sample allocation recovers DS-level training
signal at a fraction of its cost, and whether it stacks with predictive
selection. Related work is deferred to App.~\ref{app:related}.

\subsection{Setup}
\label{sec:exp:setup}
\paragraph{Models, data, and compute.}
All runs use a single GPU with R1-Distill-Qwen-1.5B and
Qwen2.5-3B~\citep{guo2025deepseekr1,yang2024qwen25}. We train on the
MATH~\citep{hendrycks2021math} and Countdown~\citep{pan2025tinyzero} splits
with GRPO on verl~\citep{sheng2024verl}, and evaluate pass@1 (mean over 16
samples) on AIME24, AMC23, MATH500~\citep{lightman2023verify},
Minerva~\citep{lewkowycz2022minerva}, and
OlympiadBench~\citep{he2024olympiadbench}. Full hyper-parameters and protocol
are in App.~\ref{app:setup}.

\paragraph{Baselines.}
We compare \emph{Uniform} (fixed-$k$ GRPO), \emph{History Resampling}
(HR)~\citep{zhang2025srpo}, \emph{DPS}~\citep{mao2026dps}, and \emph{DS
(oracle)}~\citep{yu2025dapo}, plus \method{} and the composition
\method{}$+$DPS (DPS orders the draw; \method{} verifies in-sample). Unless
noted, $B{=}\saraBatchB$, $k{=}\saraGroupK$, $\ninit{=}\saraProbe$,
$\taulow{=}\saraTau$, and every method is trained for the same
$\saraSteps{}$ optimizer steps; Uniform / HR / DPS spend exactly $Bk$ rollouts
per step, while DS and \method{} keep sampling until $B$ effective groups are
filled. We report pass@1, total rollouts, and tokens to the reported checkpoint.

\begin{table}[t]
\centering
\caption{\textbf{Main results} (pass@1, avg.\ over 16 samples).
\method{} matches DPS at \saraRollSave{} fewer rollouts than DS;
\method{}$+$DPS is most accurate at \saraCompRollSave{} fewer rollouts than DS.
\textbf{Bold}/\underline{underline}: best/second among finetuned methods.
Roll.\ (M)\,/\,Tok.\ (B): totals to the reported checkpoint.}
\label{tab:main}
\resizebox{\textwidth}{!}{
\begin{tabular}{lcccccccc}
\toprule
Method & AIME24 & AMC23 & MATH500 & Minerva & Olympiad & Avg.\,$\uparrow$ & Roll.\,(M)\,$\downarrow$ & Tok.\,(B)\,$\downarrow$ \\
\midrule
\,1.5B (base) & 18.3 & 51.7 & 76.6 & 23.8 & 35.3 & 41.1 & -- & -- \\
+Uniform & 27.0 & 58.0 & 79.9 & 26.4 & 41.3 & 46.5 & 0.15 & 0.33 \\
+HR & 29.4 & 60.3 & 80.9 & 27.2 & 42.9 & 48.2 & 0.15 & 0.45 \\
+DPS & 37.3 & 66.1 & 83.9 & 29.8 & 48.5 & 53.1 & 0.15 & 0.34 \\
+DS (oracle) & \underline{38.8} & \underline{67.1} & \underline{84.3} & \underline{30.3} & \underline{49.5} & \underline{54.0} & 0.63 & 1.37 \\
+\method (ours) & 37.6 & 66.3 & 84.0 & 29.9 & 48.8 & 53.3 & 0.49 & 1.05 \\
+\method+DPS (ours) & \textbf{40.1} & \textbf{68.2} & \textbf{85.0} & \textbf{30.7} & \textbf{50.6} & \textbf{54.9} & 0.21 & 0.46 \\
\midrule
\,3B (base) & 28.0 & 60.1 & 80.2 & 29.0 & 41.0 & 47.7 & -- & -- \\
+Uniform & 35.8 & 67.0 & 84.1 & 31.9 & 46.3 & 53.0 & 0.15 & 0.33 \\
+HR & 38.6 & 69.9 & 85.5 & 33.5 & 48.5 & 55.2 & 0.15 & 0.43 \\
+DPS & 44.4 & 75.4 & 88.6 & 35.9 & 52.5 & 59.4 & 0.15 & 0.33 \\
+DS (oracle) & \underline{45.7} & \underline{76.7} & \underline{89.3} & \underline{36.4} & \underline{53.7} & \underline{60.3} & 0.63 & 1.24 \\
+\method (ours) & 44.5 & 75.7 & 88.7 & 36.1 & 52.8 & 59.5 & 0.48 & 0.95 \\
+\method+DPS (ours) & \textbf{46.6} & \textbf{77.5} & \textbf{89.8} & \textbf{37.1} & \textbf{54.3} & \textbf{61.1} & 0.21 & 0.45 \\
\bottomrule
\end{tabular}
}
\end{table}

\subsection{Main Results}
\label{sec:exp:main}
\begin{figure}[t]
\centering
\includegraphics[width=\textwidth]{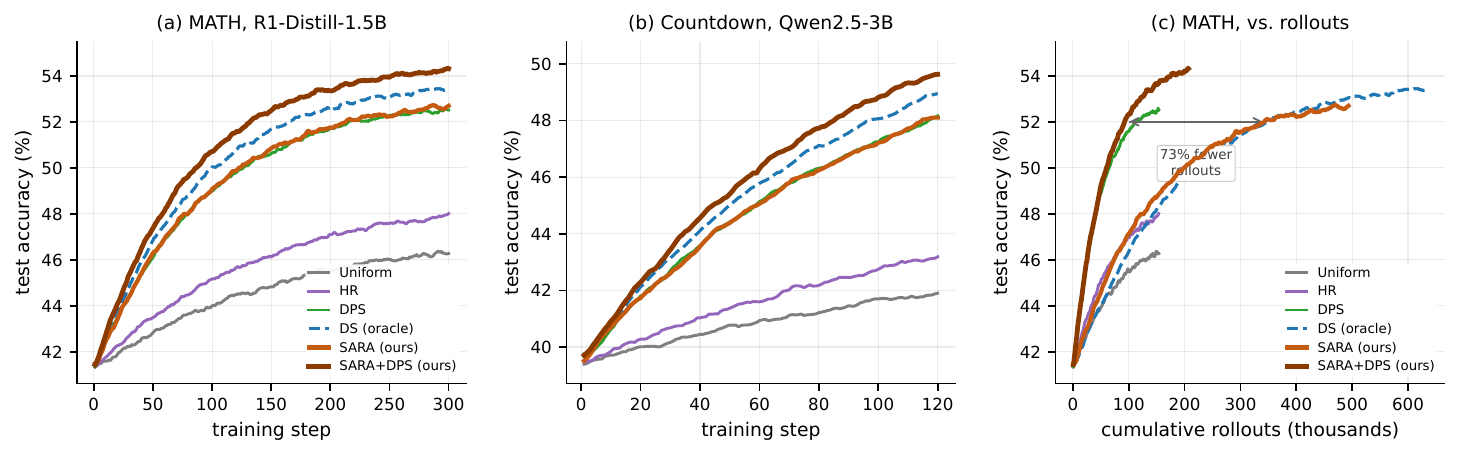}
\caption{\textbf{Accuracy and cost over training.}
(a,~b)~Pass@1 vs.\ optimizer step on MATH (1.5B) and Countdown (3B); every
method runs for the same number of steps.
(c)~MATH vs.\ cumulative rollouts: \method{}$+$DPS reaches above-DS accuracy
with \saraCompRollSave{} fewer rollouts.}
\label{fig:training}
\end{figure}

Table~\ref{tab:main} and \Figref{fig:training} summarize the comparison.
Uniform improves over the base model but plateaus: most of its batch is
saturated (effective fraction $\approx\!\saraEffFrac{}$). DPS recovers much of
the missing signal at Uniform cost, but its batch remains only as clean as its
forecast. DS guarantees a clean batch at roughly $4\times$ Uniform rollouts.
\method{} reaches DPS-level accuracy while spending \saraRollSave{} fewer
rollouts than DS, by abandoning saturated groups early and recycling budget
toward prompts that still carry within-group variance. Composing the two is
strongest: DPS supplies a cheap ordering over the prompt stream, and \method{}
supplies the in-sample guarantee, so \method{}$+$DPS exceeds DS on accuracy at
\saraCompRollSave{} fewer rollouts.

\subsection{Mechanism and Ablations}
\label{sec:exp:mech}
\Figref{fig:mechanism}a confirms early decidability: most groups are decided
within $2$--$4$ rollouts, and the verdicts agree with the oracle
effective/saturated label. At the budget that lets DS assemble a clean batch,
\method{} reaches the same clean batch earlier, using
$\sim$\saraRollSave{}/\saraTokSave{} fewer rollouts/tokens
(\Figref{fig:savings}). Token savings are amplified because abandoned all-fail
traces are the longest and are cut after a short probe.

\Figref{fig:motivation}b isolates prediction versus verification. DPS lifts the
effective fraction from Uniform's $\approx\!\saraEffFrac{}$ into the
$0.8$--$0.9$ range but never reaches a fully clean batch: a wrong forecast still
spends a full group. \method{} holds the effective fraction near $1$ throughout
by reading current outcomes. \method{}$+$DPS removes the residual gap by letting
prediction order the queue while verification decides.

\begin{figure}[t]
\centering
\includegraphics[width=\textwidth]{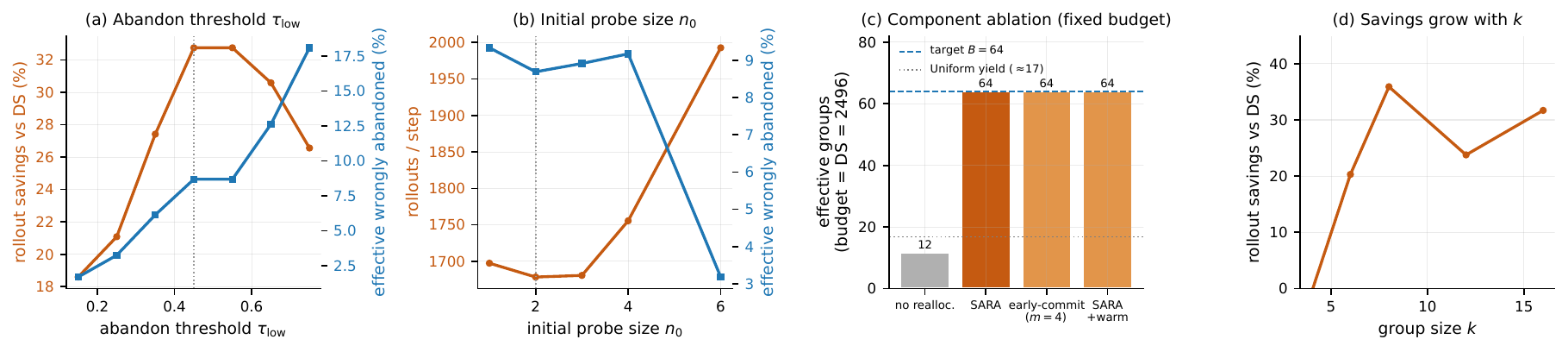}
\caption{\textbf{Ablations.} (a) abandon threshold $\taulow$; (b) probe size
$\ninit$; (c) component ablation (effective groups under a fixed budget equal
to DS's per-step cost); (d) savings vs.\ DS grow with group size $k$.}
\label{fig:ablation}
\end{figure}

\Figref{fig:ablation} ablates the design (details in App.~\ref{app:extra}).
Larger $\taulow$ abandons earlier but increases wrong abandonments, matching
Thm.~\ref{thm:reliability}; $\taulow{=}\saraTau$ saves $\sim$\saraRollSave{} of
rollouts at $\saraLostRate{}$ abandonment loss. Results are robust to
$\ninit\in\{1,\dots,4\}$. Under a fixed budget equal to DS's per-step cost,
\method{} fills $B$ effective groups while Uniform recovers only $\Phi B$;
disabling reallocation collapses the yield. Savings over DS rise with $k$, as
predicted by Thm.~\ref{thm:savings}.

\subsection{Compatibility}
\label{sec:exp:compat}
\method{} acts on rollout collection, upstream of the optimizer, so it applies to
any group-based RL algorithm. Table~\ref{tab:compat} shows consistent gains when
\method{} replaces uniform collection under GRPO, PPO, RLOO, and
Reinforce++~\citep{schulman2017ppo,hu2025reinforcepp}. The same composition with
DPS remains complementary across these recipes (App.~\ref{app:extra}).

\begin{table}[t]
\centering
\caption{\textbf{\method{} is algorithm-agnostic} (Countdown, Qwen2.5-3B, pass@1
on CD-34 and the harder CD-4). \method{} improves every base RL algorithm at
matched effective-batch size.}
\label{tab:compat}
\begin{tabular}{lcccc}
\toprule
& \multicolumn{2}{c}{Uniform} & \multicolumn{2}{c}{+\,\method} \\
\cmidrule(lr){2-3}\cmidrule(lr){4-5}
RL algorithm & CD-34 & CD-4 & CD-34 & CD-4 \\
\midrule
PPO ($k{=}8$) & 69.6 & 40.3 & \textbf{72.9} & \textbf{44.9} \\
GRPO ($k{=}8$) & 70.0 & 40.7 & \textbf{73.3} & \textbf{45.3} \\
RLOO ($k{=}8$) & 69.3 & 39.8 & \textbf{72.6} & \textbf{44.4} \\
Reinforce++ ($k{=}8$) & 69.1 & 39.3 & \textbf{72.4} & \textbf{43.9} \\
\bottomrule
\end{tabular}

\end{table}

\section{Conclusion}
\label{sec:conclusion}
We reframed per-step RLVR rollout collection as sequential allocation and
introduced \method{}, which decides commit / abandon / continue from a group's
own rollouts via a Beta--Binomial predictor and a two-threshold stopping rule.
The allocator needs no prediction rollouts, drops into GRPO-style pipelines, and
stacks with predictive selection. Empirically, \method{} matches DPS-level
accuracy at a fraction of DS cost, and \method{}$+$DPS is both most accurate and
cheapest among methods that fill every batch. Theory links abandonment
reliability, expected savings, and gradient signal to the same effectiveness
predictor that drives the algorithm.
\textbf{Limitations.}
\method{} assumes binary verifiable rewards and i.i.d.\ rollouts within a group;
extending the predictor to continuous rewards and correlated tree rollouts
(App.~\ref{app:discussion}), and validating at larger scale, remain open.

\subsubsection*{Reproducibility Statement}
Algorithmic details appear in \Secref{sec:method} and Algorithm~\ref{alg:sara},
with proofs in App.~\ref{app:proofs}, implementation notes in App.~\ref{app:impl},
and the full experimental protocol in App.~\ref{app:setup}. We will release code
and scripts to reproduce all figures and tables upon publication.

\subsubsection*{Ethics Statement}
\method{} is a training-efficiency method for RLVR; by reducing the rollouts and
tokens needed per update it lowers the energy and monetary cost of reasoning-LLM
post-training. It inherits the general risks of more capable reasoning models
and does not introduce new data or human-subject concerns. Our use of large
language models in preparing this paper is disclosed in App.~\ref{app:llm}.

\bibliography{custom}

@article{guo2025deepseekr1,
  title={DeepSeek-R1: Incentivizing Reasoning Capability in {LLMs} via Reinforcement Learning},
  author={Guo, Daya and Yang, Dejian and Zhang, Haowei and Song, Junxiao and Zhang, Ruoyu and Xu, Runxin and Zhu, Qihao and Ma, Shirong and Wang, Peiyi and Bi, Xiao and others},
  journal={arXiv preprint arXiv:2501.12948},
  year={2025}
}

@article{jaech2024o1,
  title={OpenAI o1 System Card},
  author={Jaech, Aaron and Kalai, Adam and Lerer, Adam and Richardson, Adam and El-Kishky, Ahmed and Low, Aiden and Helyar, Alec and Madry, Aleksander and Beutel, Alex and Carney, Alex and others},
  journal={arXiv preprint arXiv:2412.16720},
  year={2024}
}

@article{kimi2025k15,
  title={Kimi k1.5: Scaling Reinforcement Learning with {LLMs}},
  author={Kimi Team and Du, Angang and Gao, Bofei and Xing, Bowei and Jiang, Changjiu and Chen, Cheng and Li, Cheng and Xiao, Chenjun and Du, Chenzhuang and Liao, Chonghua and others},
  journal={arXiv preprint arXiv:2501.12599},
  year={2025}
}

@article{shao2024deepseekmath,
  title={DeepSeekMath: Pushing the Limits of Mathematical Reasoning in Open Language Models},
  author={Shao, Zhihong and Wang, Peiyi and Zhu, Qihao and Xu, Runxin and Song, Junxiao and Bi, Xiao and Zhang, Haowei and Zhang, Mingchuan and Li, Y. K. and Wu, Y. and others},
  journal={arXiv preprint arXiv:2402.03300},
  year={2024}
}

@article{schulman2017ppo,
  title={Proximal Policy Optimization Algorithms},
  author={Schulman, John and Wolski, Filip and Dhariwal, Prafulla and Radford, Alec and Klimov, Oleg},
  journal={arXiv preprint arXiv:1707.06347},
  year={2017}
}

@article{lightman2023verify,
  title={Let's Verify Step by Step},
  author={Lightman, Hunter and Kosaraju, Vineet and Burda, Yura and Edwards, Harrison and Baker, Bowen and Lee, Teddy and Leike, Jan and Schulman, John and Sutskever, Ilya and Cobbe, Karl},
  journal={arXiv preprint arXiv:2305.20050},
  year={2023}
}

@article{hu2025reinforcepp,
  title={{REINFORCE++}: Stabilizing Critic-Free Policy Optimization with Global Advantage Normalization},
  author={Hu, Jian},
  journal={arXiv preprint arXiv:2501.03262},
  year={2025}
}

@article{yu2025dapo,
  title={DAPO: An Open-Source {LLM} Reinforcement Learning System at Scale},
  author={Yu, Qiying and Zhang, Zheng and Zhu, Ruofei and Yuan, Yufeng and Zuo, Xiaochen and Yue, Yu and Fan, Tiantian and Liu, Gaohong and Liu, Lingjun and Liu, Xin and others},
  journal={arXiv preprint arXiv:2503.14476},
  year={2025}
}

@article{bae2025online,
  title={Online Difficulty Filtering for Reasoning Oriented Reinforcement Learning},
  author={Bae, Sanghwan and Hong, Jiwoo and Lee, Min Young and Kim, Hanbyul and Nam, JeongYeon and Kwak, Donghyun},
  journal={arXiv preprint arXiv:2504.03380},
  year={2025}
}

@article{chen2025sec,
  title={Self-Evolving Curriculum for {LLM} Reasoning},
  author={Chen, Xiaoyin and Lu, Jiarui and Kim, Minsu and Zhang, Dinghuai and Tang, Jian and Pich{\'e}, Alexandre and Gontier, Nicolas and Bengio, Yoshua and Kamalloo, Ehsan},
  journal={arXiv preprint arXiv:2505.14970},
  year={2025}
}

@article{zheng2025selective,
  title={Act Only When It Pays: Efficient Reinforcement Learning for {LLM} Reasoning via Selective Rollouts},
  author={Zheng, Haizhong and Zhou, Yang and Bartoldson, Brian R and Kailkhura, Bhavya and Lai, Fan and Zhao, Jiawei and Chen, Beidi},
  journal={arXiv preprint arXiv:2506.02177},
  year={2025}
}

@article{lin2025cppo,
  title={CPPO: Accelerating the Training of Group Relative Policy Optimization-Based Reasoning Models},
  author={Lin, Zhihang and Lin, Mingbao and Xie, Yuan and Ji, Rongrong},
  journal={arXiv preprint arXiv:2503.22342},
  year={2025}
}

@article{zhang2025srpo,
  title={SRPO: A Cross-Domain Implementation of Large-Scale Reinforcement Learning on {LLM}},
  author={Zhang, Xiaojiang and Wang, Jinghui and Cheng, Zifei and Zhuang, Wenhao and Lin, Zheng and Zhang, Minglei and Wang, Shaojie and Cui, Yinghan and Wang, Chao and Peng, Junyi and others},
  journal={arXiv preprint arXiv:2504.14286},
  year={2025}
}

@inproceedings{mao2026dps,
  title={Dynamics-Predictive Sampling for Active {RL} Finetuning of Large Reasoning Models},
  author={Mao, Yixiu and Qu, Yun and Wang, Qi and Zou, Heming and Ji, Xiangyang},
  booktitle={International Conference on Learning Representations ({ICLR})},
  year={2026}
}

@inproceedings{qu2026gps,
  title={Small Generalizable Prompt Predictive Models Can Steer Efficient {RL} Post-Training of Large Reasoning Models},
  author={Qu, Yun and Wang, Qi and Mao, Yixiu and Zou, Heming and Jiang, Yuhang and Liu, Weijie and Bai, Clive and Yang, Kai and Chen, Yangkun and Yang, Saiyong and Ji, Xiangyang},
  booktitle={International Conference on Machine Learning ({ICML})},
  year={2026}
}

@article{zou2026trace,
  title={TRACE: A Unified Rollout Budget Allocation Framework for Efficient Agentic Reinforcement Learning},
  author={Zou, Heming and Wang, Qi and Qu, Yun and Jiang, Yuhang and Cai, Lizhou and Mao, Yixiu and Peng, Ru and Xu, Xin and Liu, Weijie and Yang, Kai and Yang, Saiyong and Ji, Xiangyang},
  journal={arXiv preprint arXiv:2606.11119},
  year={2026}
}

@article{liu2025prorl,
  title={ProRL: Prolonged Reinforcement Learning Expands Reasoning Boundaries in Large Language Models},
  author={Liu, Mingjie and Diao, Shizhe and Lu, Ximing and Hu, Jian and Dong, Xin and Choi, Yejin and Kautz, Jan and Dong, Yi},
  journal={arXiv preprint arXiv:2505.24864},
  year={2025}
}

@article{li2025limr,
  title={LIMR: Less is More for {RL} Scaling},
  author={Li, Xuefeng and Zou, Haoyang and Liu, Pengfei},
  journal={arXiv preprint arXiv:2502.11886},
  year={2025}
}

@article{liu2025drgrpo,
  title={Understanding R1-Zero-Like Training: A Critical Perspective},
  author={Liu, Zichen and Chen, Changyu and Li, Wenjun and Qi, Penghui and Pang, Tianyu and Du, Chao and Lee, Wee Sun and Lin, Min},
  journal={arXiv preprint arXiv:2503.20783},
  year={2025}
}

@article{wang2025oneshot,
  title={Reinforcement Learning for Reasoning in Large Language Models with One Training Example},
  author={Wang, Yiping and Yang, Qing and Zeng, Zhiyuan and Ren, Liliang and Liu, Liyuan and Peng, Baolin and Cheng, Hao and He, Xuehai and Wang, Kuan and Gao, Jianfeng and others},
  journal={arXiv preprint arXiv:2504.20571},
  year={2025}
}

@article{fatemi2025concise,
  title={Concise Reasoning via Reinforcement Learning},
  author={Fatemi, Mehdi and Rafiee, Banafsheh and Tang, Mingjie and Talamadupula, Kartik},
  journal={arXiv preprint arXiv:2504.05185},
  year={2025}
}

@article{aggarwal2025l1,
  title={L1: Controlling How Long A Reasoning Model Thinks With Reinforcement Learning},
  author={Aggarwal, Pranjal and Welleck, Sean},
  journal={arXiv preprint arXiv:2503.04697},
  year={2025}
}

@article{hou2025thinkprune,
  title={ThinkPrune: Pruning Long Chain-of-Thought of {LLMs} via Reinforcement Learning},
  author={Hou, Bairu and Zhang, Yang and Ji, Jiabao and Liu, Yujian and Qian, Kaizhi and Andreas, Jacob and Chang, Shiyu},
  journal={arXiv preprint arXiv:2504.01296},
  year={2025}
}

@article{sui2025overthinking,
  title={Stop Overthinking: A Survey on Efficient Reasoning for Large Language Models},
  author={Sui, Yang and Chuang, Yu-Neng and Wang, Guanchu and Zhang, Jiamu and Zhang, Tianyi and Yuan, Jiayi and Liu, Hongyi and Wen, Andrew and Zhong, Shaochen and Chen, Hanjie and others},
  journal={arXiv preprint arXiv:2503.16419},
  year={2025}
}

@article{chen2025longcot,
  title={Towards Reasoning Era: A Survey of Long Chain-of-Thought for Reasoning Large Language Models},
  author={Chen, Qiguang and Qin, Libo and Liu, Jinhao and Peng, Dengyun and Guan, Jiannan and Wang, Peng and Hu, Mengkang and Zhou, Yuhang and Gao, Te and Che, Wanxiang},
  journal={arXiv preprint arXiv:2503.09567},
  year={2025}
}

@article{ye2025limo,
  title={LIMO: Less is More for Reasoning},
  author={Ye, Yixin and Huang, Zhen and Xiao, Yang and Chern, Ethan and Xia, Shijie and Liu, Pengfei},
  journal={arXiv preprint arXiv:2502.03387},
  year={2025}
}

@article{wald1945sequential,
  title={Sequential Tests of Statistical Hypotheses},
  author={Wald, Abraham},
  journal={The Annals of Mathematical Statistics},
  volume={16},
  number={2},
  pages={117--186},
  year={1945}
}

@article{gittins1979bandit,
  title={Bandit Processes and Dynamic Allocation Indices},
  author={Gittins, John C},
  journal={Journal of the Royal Statistical Society: Series B (Methodological)},
  volume={41},
  number={2},
  pages={148--164},
  year={1979}
}

@article{robbins1952sequential,
  title={Some Aspects of the Sequential Design of Experiments},
  author={Robbins, Herbert},
  journal={Bulletin of the American Mathematical Society},
  volume={58},
  number={5},
  pages={527--535},
  year={1952}
}

@article{thompson1933likelihood,
  title={On the Likelihood that One Unknown Probability Exceeds Another in View of the Evidence of Two Samples},
  author={Thompson, William R},
  journal={Biometrika},
  volume={25},
  number={3-4},
  pages={285--294},
  year={1933}
}

@article{berry1972bernoulli,
  title={A Bernoulli Two-Armed Bandit},
  author={Berry, Donald A},
  journal={The Annals of Mathematical Statistics},
  volume={43},
  number={3},
  pages={871--897},
  year={1972}
}

@article{russo2014posterior,
  title={Learning to Optimize via Posterior Sampling},
  author={Russo, Daniel and Van Roy, Benjamin},
  journal={Mathematics of Operations Research},
  volume={39},
  number={4},
  pages={1221--1243},
  year={2014}
}

@article{auer2002finite,
  title={Finite-Time Analysis of the Multiarmed Bandit Problem},
  author={Auer, Peter and Cesa-Bianchi, Nicol{\`o} and Fischer, Paul},
  journal={Machine Learning},
  volume={47},
  number={2},
  pages={235--256},
  year={2002}
}

@article{hendrycks2021math,
  title={Measuring Mathematical Problem Solving With the {MATH} Dataset},
  author={Hendrycks, Dan and Burns, Collin and Kadavath, Saurav and Arora, Akul and Basart, Steven and Tang, Eric and Song, Dawn and Steinhardt, Jacob},
  journal={arXiv preprint arXiv:2103.03874},
  year={2021}
}

@article{lewkowycz2022minerva,
  title={Solving Quantitative Reasoning Problems with Language Models},
  author={Lewkowycz, Aitor and Andreassen, Anders and Dohan, David and Dyer, Ethan and Michalewski, Henryk and Ramasesh, Vinay and Slone, Ambrose and Anil, Cem and Schlag, Imanol and Gutman-Solo, Theo and others},
  journal={Advances in Neural Information Processing Systems},
  volume={35},
  pages={3843--3857},
  year={2022}
}

@article{he2024olympiadbench,
  title={OlympiadBench: A Challenging Benchmark for Promoting {AGI} with Olympiad-Level Bilingual Multimodal Scientific Problems},
  author={He, Chaoqun and Luo, Renjie and Bai, Yuzhuo and Hu, Shengding and Thai, Zhen Leng and Shen, Junhao and Hu, Jinyi and Han, Xu and Huang, Yujie and Zhang, Yuxiang and others},
  journal={arXiv preprint arXiv:2402.14008},
  year={2024}
}

@misc{pan2025tinyzero,
  title={TinyZero},
  author={Pan, Jiayi and Zhang, Junjie and Wang, Xingyao and Yuan, Lifan and Peng, Hao and Suhr, Alane},
  year={2025},
  howpublished={\url{https://github.com/Jiayi-Pan/TinyZero}}
}

@article{sheng2024verl,
  title={HybridFlow: A Flexible and Efficient {RLHF} Framework},
  author={Sheng, Guangming and Zhang, Chi and Ye, Zilingfeng and Wu, Xibin and Zhang, Wang and Zhang, Ru and Peng, Yanghua and Lin, Haibin and Wu, Chuan},
  journal={arXiv preprint arXiv:2409.19256},
  year={2024}
}

@article{yang2024qwen25,
  title={Qwen2.5 Technical Report},
  author={Yang, An and Yang, Baosong and Zhang, Beichen and Hui, Binyuan and Zheng, Bo and Yu, Bowen and Li, Chengyuan and Liu, Dayiheng and Huang, Fei and Wei, Haoran and others},
  journal={arXiv preprint arXiv:2412.15115},
  year={2024}
}
\bibliographystyle{iclr2026_conference}

\appendix
\newpage
\section{Use of Large Language Models}
\label{app:llm}
Large language models were used solely as general-purpose assistive tools during
manuscript preparation: polishing wording, checking LaTeX, and suggesting
references that the authors subsequently verified against primary sources. No
LLM was used to generate experimental results, proofs, or claims; all
theoretical statements and their proofs (App.~\ref{app:proofs}) were written and
verified by the authors. The method itself (\method{}) concerns the RL training
of LLMs but does not use an LLM as part of the research methodology beyond the
standard RLVR pipeline it studies.

\section{Related Work}
\label{app:related}
We organize prior work along the prediction-versus-verification axis of
\Secref{sec:intro} and place \method{} between its two extremes.

\paragraph{RLVR and group-relative optimization.}
RLVR scales reasoning by optimizing a policy against an automatically verifiable
reward~\citep{guo2025deepseekr1,jaech2024o1,kimi2025k15}. GRPO
\citep{shao2024deepseekmath} replaced PPO's~\citep{schulman2017ppo} value network
with a group-normalized advantage; many variants tune stability, bias, and
length~\citep{liu2025drgrpo,hu2025reinforcepp,yu2025dapo,liu2025prorl}. The
recurring principle is that the \emph{within-group reward variance} is the source
of signal (exactly what saturated groups lack), which makes assembling
effective groups the central efficiency lever and motivates \method{}.

\paragraph{Sample-efficient RLVR.}
Efforts to spend the rollout budget better fall into two main groups.
\emph{(i) Prompt selection} chooses which prompts to train on: offline curation
ranks prompts by static difficulty or diversity~\citep{li2025limr,ye2025limo,
wang2025oneshot}; online \emph{evaluate-then-filter} methods (dynamic sampling
in DAPO~\citep{yu2025dapo} and online difficulty filtering~\citep{bae2025online})
oversample a candidate pool, generate full groups, and discard saturated
ones; and \emph{predict-then-select} methods forecast difficulty \emph{before}
sampling~\citep{mao2026dps,qu2026gps,chen2025sec}.
\emph{(ii) Rollout allocation} sets how many rollouts each prompt or trajectory
prefix receives~\citep{zou2026trace} and prunes low-value
generations~\citep{zheng2025selective,lin2025cppo}.
These all commit budget \emph{before} a group's
own rollouts are observed (evaluate-then-filter even pays full groups for the
prompts it discards). \method{} instead decides \emph{during} a group's own
rollouts with an in-sample sequential verdict: it needs no auxiliary evaluation
passes, is robust to the prediction error that limits predict-then-select, and
composes with prompt selection (our \method{}$+$DPS results in
\Secref{sec:exp:compat} confirm the two axes stack). The closest point of
comparison is selective rollouts~\citep{zheng2025selective}, which prune by a
heuristic cutoff; \method{} replaces the heuristic with a closed-form
sequential-testing rule carrying reliability, savings, and yield guarantees, and
adds explicit budget \emph{reallocation}. \method{} is also compatible with
tree rollouts (App.~\ref{app:discussion}).

\paragraph{Efficient reasoning and length.}
A parallel line reduces the \emph{token} cost of each rollout by controlling
reasoning length via RL penalties or pruning~\citep{aggarwal2025l1,
hou2025thinkprune,fatemi2025concise}, surveyed by \citet{sui2025overthinking,
chen2025longcot}. \method{} targets the orthogonal \emph{rollout-count} axis and
composes with length control; it already exploits the same observation that
long all-fail traces are especially wasteful (\Figref{fig:savings}b).

\paragraph{Sequential analysis, bandits, and optimal stopping.}
\method{}'s formulation draws on classical sequential analysis (Wald's SPRT and
the sequential design of
experiments~\citep{wald1945sequential,robbins1952sequential}) and posterior
sampling~\citep{thompson1933likelihood,berry1972bernoulli,russo2014posterior,
auer2002finite}. Casting one RLVR step as a budget-constrained restless bandit
connects \method{} to Gittins-style index policies~\citep{gittins1979bandit}, of
which \method{} is the tractable myopic instance (App.~\ref{app:proofs:gittins}).
To our knowledge, this is the first application of sequential hypothesis testing
to RLVR rollout budgeting.

\section{Proofs and Theoretical Analysis}
\label{app:proofs}

Throughout, a prompt has latent success rate $\gamma\in[0,1]$, a group has target
size $k$, and rollouts are conditionally i.i.d.\ $\Bernoulli(\gamma)$ given
$\gamma$. We use a Beta prior $\gamma\sim\Beta(\alpha_0,\beta_0)$. After $n$
rollouts with $s$ successes the posterior is
$\gamma\mid(n,s)\sim\Beta(\alpha_0+s,\beta_0+n-s)$. Recall the Beta-function
identity $\E_{\gamma\sim\Beta(a,b)}[\gamma^{u}(1-\gamma)^{v}] =
\Betaf(a+u,b+v)/\Betaf(a,b)$ for integers $u,v\ge0$.

\subsection{Proof of Proposition~\ref{prop:predictor} (closed-form predictor)}
A completed group of size $k$ is effective iff $1\le S_k\le k-1$. If the observed
prefix is already mixed ($s\ge1$ and $n-s\ge1$) then $S_k\ge s\ge1$ and
$k-S_k\ge (n-s)\ge1$, so the group is effective with probability $1$. Otherwise
the prefix is all-fail ($s=0$) or all-pass ($s=n$); consider $s=0$ (the other
case is symmetric by $\gamma\leftrightarrow1-\gamma$). Let $r=k-n$ be the
remaining rollouts. The group fails to be effective iff all $r$ remaining
rollouts also fail (then $S_k=0$); it cannot be all-pass since the prefix already
contains $n$ failures. Hence
\begin{align}
\peff(n,0)
&= 1-\Pr[\text{remaining $r$ all fail}\mid n,0]
= 1-\E_{\gamma\sim\Beta(\alpha_0,\beta_0+n)}\!\big[(1-\gamma)^{r}\big]\nonumber\\
&= 1-\frac{\Betaf(\alpha_0,\,\beta_0+n+r)}{\Betaf(\alpha_0,\,\beta_0+n)},
\end{align}
using the identity with $u=0,v=r$ and posterior parameters $(\alpha_0,\beta_0+n)$.
With a uniform prior $\alpha_0=\beta_0=1$ and $\Betaf(1,m)=1/m$,
$\peff(n,0)=1-\frac{(\beta_0+n)}{(\beta_0+n+r)}=1-\frac{1+n}{1+k}=\frac{k-n}{k+1}$,
since $\beta_0+n+r=1+n+(k-n)=1+k$. The all-pass case is identical with
$\alpha_0\leftrightarrow\beta_0$ and $s=n$. \qed

\subsection{Proof of Proposition~\ref{prop:monotone} (irreversibility, monotonicity)}
\emph{Irreversibility.} If a prefix is mixed, then $S_k\ge s\ge1$ and
$k-S_k\ge n-s\ge1$ for every completion, so the group remains effective; the
event ``effective'' is absorbing along any rollout sequence.
\emph{Monotonicity.} Along an all-fail prefix, by Proposition~\ref{prop:predictor}
$\peff(n,0)=1-\Betaf(\alpha_0,\beta_0+k)/\Betaf(\alpha_0,\beta_0+n)$ because
$\beta_0+n+r=\beta_0+k$ is constant in $n$. The map $x\mapsto\Betaf(\alpha_0,x)
=\Gamma(\alpha_0)\Gamma(x)/\Gamma(\alpha_0+x)$ is strictly decreasing in $x>0$
(its logarithmic derivative is $\psi(x)-\psi(\alpha_0+x)<0$, where $\psi$ is the
digamma function). Hence $\Betaf(\alpha_0,\beta_0+n)$ strictly decreases in $n$,
so the ratio $\Betaf(\alpha_0,\beta_0+k)/\Betaf(\alpha_0,\beta_0+n)$ strictly
\emph{increases}, and $\peff(n,0)$ strictly decreases. Therefore each same-outcome
run crosses the threshold $\taulow$ at a unique time, and every prompt is decided
(committed when it first becomes mixed, or abandoned at threshold crossing) at a
well-defined stopping time $\le k$. The fraction of groups decided by round $n$
is the CDF of this stopping time, hence nondecreasing and $\to1$ at $n=k$. \qed

\subsection{Proof of Theorem~\ref{thm:reliability} (abandonment reliability)}
\method{} abandons a prompt only from an all-same prefix with
$\peff(n,s)<\taulow$. By Proposition~\ref{prop:predictor}, $\peff(n,s)$ is exactly
the posterior probability that the completed group would be effective given the
observed prefix. Thus, conditioned on the data at the abandonment time, the
probability that the abandoned group ``would have been effective'' equals
$\peff<\taulow$. Let $\mathcal{B}$ be the (random) set of abandoned prompts in a
step. The number of lost effective groups is
$L=\sum_{q\in\mathcal{B}}\mathbf 1[q\text{ would be effective}]$, so by the tower
rule $\E[L]=\E\big[\sum_{q\in\mathcal{B}}\peff_q\big]\le\taulow\,\E|\mathcal{B}|$.
As $\taulow\to0$, abandonment requires $\peff=0$, i.e.\ $n=k$ with an all-same
group (a genuinely saturated group), so no effective group is ever lost:
\method{} is lossless in the limit. Finally, the \emph{commit} side is exact---a
group is committed only after a mixed outcome is observed, which by
Proposition~\ref{prop:monotone} guarantees effectiveness---so the only error mode
is the one bounded above. Viewing the all-same run as a one-sided sequential
probability ratio test of ``saturated'' against ``mixed'', $\taulow$ is exactly
the bound on its type-II error, while the type-I error is zero. \qed

\subsection{Proof of Theorem~\ref{thm:savings} (expected rollout savings)}
Fix the uniform prior; the abandonment time along a same-outcome run is the first
$n$ with $\peff(n,0)=\frac{k-n}{k+1}<\taulow$, i.e.\
$n_a=k-\lfloor\taulow(k+1)\rfloor$ (capped at $k$). Let $T_{\mathrm{mix}}$ be the
first round at which both outcomes have appeared. For a $\Bernoulli(\gamma)$
stream, $\Pr[T_{\mathrm{mix}}>n]=\gamma^{n}+(1-\gamma)^{n}$ for $n\ge1$ and $=1$
for $n=0$. \method{} stops a prompt at $N=\min(T_{\mathrm{mix}},\,n_a)$ (commit if
$T_{\mathrm{mix}}\le n_a$, else abandon), so
\begin{equation}
\E[N(\gamma)]=\sum_{n=0}^{n_a-1}\Pr[N>n]=1+\sum_{n=1}^{n_a-1}\big(\gamma^{n}+(1-\gamma)^{n}\big)\;\le\;n_a\;\le\;k,
\label{eq:EN}
\end{equation}
with equality $E[N]=n_a$ as $\gamma\to0$ or $1$ (always same-outcome until
abandonment) and $\E[N]\to2$ as $\gamma\to\tfrac12$ (mixes almost immediately).
Dynamic sampling generates a \emph{full} group of $k$ rollouts for every prompt
it evaluates (kept or discarded), i.e.\ a per-prompt cost of exactly $k$. Since
both schemes must evaluate, in expectation, $B/\Phi$ prompts to obtain $B$
effective groups (each prompt is effective with probability $\Phi=\E_\gamma
\Phi(\gamma)$), the expected per-step costs are
\begin{equation}
\mathrm{cost}_{\textsc{ds}}=\frac{B}{\Phi}\,k,\qquad
\mathrm{cost}_{\method{}}=\frac{B}{\Phi}\,\E_\gamma[N(\gamma)]
\;\le\;\frac{B}{\Phi}\,k=\mathrm{cost}_{\textsc{ds}},
\end{equation}
with strict inequality whenever $\Pr[\gamma\notin\{$values that mix before
$n_a\}]>0$, i.e.\ whenever any prompt is ever abandoned or committed early. The
relative saving $1-\E_\gamma[N(\gamma)]/k$ increases with $k$: $n_a/k\to
1-\taulow$ while early-commit caps effective groups at the commit floor
independent of $k$, so for large $k$ both terms in $\E[N]/k\to$ a constant
$<1$ shrinking the ratio. (With an informative prior the same argument yields a
geometric concentration $n_a=O(\log(1/\taulow)/\log(1/\max(\gamma,1-\gamma)))$.)
\qed

\subsection{Proof of Theorem~\ref{thm:yield} (fixed-budget yield dominance)}
Fix a budget of $\Bud$ rollouts. Uniform fixed-$k$ allocation spends them on
$\Bud/k$ full groups, yielding $Y_{\mathrm{unif}}=\Phi\cdot \Bud/k$ effective
groups in expectation. Consider \method{} run under the same budget. We use an
exchange argument. Take any saturated prompt that uniform would run to $k$
(contributing $0$ effective groups for $k$ rollouts). Under \method{} this prompt
is abandoned at $N\le n_a<k$ rollouts (Eq.~\eqref{eq:EN}); the saved $k-N>0$
rollouts are reallocated to fresh prompts, each of which is effective with
probability $\Phi>0$ and contributes a nonnegative number of effective groups in
expectation. Replacing each such saturated full group by [abandon early $+$
reallocate] therefore does not decrease the expected effective count, and
strictly increases it whenever $\Phi>0$ and at least one fresh group can be
started with the freed budget. Likewise, early-committing an effective group at
$N<k$ frees budget without losing the group. Summing these nonnegative exchanges,
$\E[Y_{\method{}}]\ge Y_{\mathrm{unif}}$, with strict inequality whenever the
success-rate distribution places mass on saturated regions (so abandonments
occur). \qed

\subsection{Proof of Theorem~\ref{thm:gradient} (effective yield bounds the gradient)}
For a saturated group the rewards are constant, so the group standard deviation
is $0$ and every normalized advantage $\hat A_i$ in Eq.~\eqref{eq:grpo} is $0$
(equivalently, the degenerate group is dropped); its contribution to
$\nabla_\vtheta\gJ$ is the zero vector. Hence
$\nabla_\vtheta\gJ=\sum_{g\in\gE}\vg_g$, where $\gE$ is the set of effective
groups and $\vg_g$ is group $g$'s gradient. Then
\begin{equation}
\E\|\nabla_\vtheta\gJ\|^2
=\E\Big\|\sum_{g\in\gE}\vg_g\Big\|^2
=\E\!\!\sum_{g,g'\in\gE}\!\!\langle\vg_g,\vg_{g'}\rangle
=\E\!\sum_{g\in\gE}\|\vg_g\|^2 + \E\!\!\sum_{g\neq g'}\!\langle\vg_g,\vg_{g'}\rangle.
\end{equation}
Assume (i) a normalized per-group energy $\E[\|\vg_g\|^2\mid g\in\gE]\ge c>0$ and
(ii) nonnegative average cross-alignment $\E\sum_{g\neq g'}\langle
\vg_g,\vg_{g'}\rangle\ge0$ (e.g.\ gradients of distinct effective groups are
conditionally uncorrelated, the standard mini-batch assumption, giving equality
to $0$). Then $\E\|\nabla_\vtheta\gJ\|^2\ge c\,\E|\gE|$. Thus maximizing the
expected number of effective groups maximizes a lower bound on the expected
squared update magnitude, which connects \method{}'s allocation objective to
optimization progress. \qed

\subsection{Optimal stopping and the index-policy view}
\label{app:proofs:gittins}
We make precise the claim that \method{} is a myopic index policy.
\begin{proposition}[\method{} as a sequential-allocation index policy]
\label{prop:gittins}
Consider the per-step problem of allocating a rollout budget across a stream of
prompts (arms), each an unknown-$\gamma$ Bernoulli process with $\Beta$ belief,
where committing an effective group yields unit reward and each rollout costs one
unit of budget. (a) Under a single-arm relaxation (process one prompt to its
stopping time before moving on), the optimal policy is a stopping rule that
continues while the expected value of one more rollout exceeds its cost; for the
effective-group objective this boundary is exactly the two-threshold rule of
Eq.~\eqref{eq:decide} with $\taulow$ set by the cost/reward ratio. (b) For the
full restless problem, the Gittins index~\citep{gittins1979bandit} gives the
optimal index policy in the discounted relaxation; \method{}'s priority by
$\peff$ is its one-step (myopic) approximation, and posterior-sampling
priority~\citep{thompson1933likelihood,russo2014posterior} is an alternative with
the same fixed points.
\end{proposition}
\emph{Sketch.} (a) The state of an arm is its posterior $\Beta(\alpha_0+s,
\beta_0+n-s)$; the value function $V(n,s)$ satisfies the Bellman optimality
equation $V=\max\{\text{stop value},\ \text{continue value}\}$. The stop value of
a mixed prefix is $1$ (an effective group in hand); the continue value of an
all-same prefix equals the discounted probability of eventually mixing, a
monotone transform of $\peff$ by Proposition~\ref{prop:monotone}; equating the
two yields a threshold on $\peff$, i.e.\ Eq.~\eqref{eq:decide}. (b) is the
standard Gittins decomposition for a bank of independent arms; the myopic index
$\peff$ is exact when arms cannot be revisited (our reallocation draws fresh
arms), and otherwise first-order optimal. \qed

\section{Algorithm and Implementation Details}
\label{app:impl}

\paragraph{Round-synchronous batched scheduling.}
A naive ``sample one rollout, decide, repeat'' loop would destroy the throughput
of batched inference engines. \method{} avoids this: each round issues a
\emph{single} batched generation call over all currently active prompts (the
initial round generates $\ninit$ rollouts per prompt; later rounds generate one
continuation per continuing prompt). Decisions, posterior updates, and
reallocation happen between rounds on CPU in $O(\text{active prompts})$ time.
Because most groups are decided within $2$--$4$ rounds
(\Figref{fig:mechanism}a), the number of synchronization points per step is
small and the per-call batch stays large, so wall-clock overhead relative to a
single monolithic generation is minor while the rollout/token \emph{count} drops
substantially.

\paragraph{Integration with verl / TRL.}
\method{} replaces only the rollout-collection stage of a GRPO trainer; the loss
and optimizer are untouched. Concretely (see \texttt{train\_grpo\_sara.py}), one
implements a single closure \texttt{rollout\_fn(prompt\_ids, count)} that maps a
list of prompts and a per-prompt count to a batched \texttt{generate} call and
returns, per rollout, the verifiable reward and token length; \method{}'s
scheduler then orchestrates the rounds and hands the optimizer a list of
effective groups. This is a $\sim$30-line change to a verl
\texttt{DataProto} generation hook or a TRL \texttt{GRPOTrainer} sampling step.

\paragraph{Variable group sizes.}
With early commit (\texttt{commit\_min}$=m$) effective groups have size in
$[m,k]$. GRPO's within-group normalization (Eq.~\ref{eq:grpo}) is well defined
for any group size $\ge2$; we pad/mask to the longest group in a micro-batch.
The default sets \texttt{commit\_min}$=k$ (fixed-size groups), so the cost
comparison with DS is not confounded by group size; early-commit
(\texttt{commit\_min}$=4$) is reported as an ablation in
\Figref{fig:ablation}c and further reduces rollouts at the cost of variable
group sizes.

\paragraph{Defaults and history warm-start.}
Defaults: $B{=}\saraBatchB$, $k{=}\saraGroupK$, $\ninit{=}\saraProbe$,
$\taulow{=}\saraTau$, $\texttt{commit\_min}{=}k$, prior $\Beta(1,1)$, safety
budget cap $6Bk$. The optional warm-start carries each prompt's posterior
across steps with an exponential decay $\lambda$ (so a prompt seen to be
saturated last step starts skewed and is abandoned after fewer probes),
mirroring the temporal discounting used by predictive
selection~\citep{mao2026dps}; it further reduces rollouts at the
cost of one scalar of state per prompt.

\paragraph{Complexity.}
Per prompt per round: one Beta update ($O(1)$) and one evaluation of
Eq.~\eqref{eq:peff} (a constant number of $\log\Gamma$ calls), plus an $O(\log
B)$ insertion into the priority structure. No extra forward/backward passes and
no auxiliary model. Memory overhead is two integers (or two floats with
warm-start) per active prompt.

\section{Detailed Experimental Setup}
\label{app:setup}

\paragraph{Models.}
We use R1-Distill-Qwen-1.5B~\citep{guo2025deepseekr1} and
Qwen2.5-3B~\citep{yang2024qwen25} (math) and Qwen2.5-3B (planning). Both fit
comfortably on a single GPU for both generation (vLLM, bf16) and the policy
update, which is the point of the resource-light design: the budget being
allocated is rollouts/tokens, not GPUs.

\paragraph{Datasets and evaluation.}
Training uses the MATH~\citep{hendrycks2021math} training split and a subset of
Countdown~\citep{pan2025tinyzero}. Math evaluation reports pass@1 (mean over 16
samples, temperature 1.0) on AIME24, AMC23,
MATH500~\citep{lightman2023verify}, Minerva~\citep{lewkowycz2022minerva}, and
OlympiadBench~\citep{he2024olympiadbench}. Planning reports pass@1 on the
in-distribution CD-34 and the harder CD-4 split. Rewards are binary verifiable
correctness; format rewards, when present, are binarized by thresholding.

\paragraph{RL configuration.}
GRPO on verl~\citep{sheng2024verl}: group size $k{=}\saraGroupK$, effective
batch $B{=}\saraBatchB$, learning rate $1{\times}10^{-6}$, KL coefficient
$1{\times}10^{-3}$, clip $\epsilon{=}0.2$, max generation length $4096$ tokens,
sampling temperature $1.0$, $\saraSteps{}$ update steps on MATH (matching the
DPS 1.5B schedule) and $120$ on Countdown. The base
GRPO/PPO/RLOO/Reinforce++ settings are shared across allocators so that only
the rollout-collection strategy differs; Uniform / HR / DPS spend exactly
$Bk$ rollouts per step, while DS and \method{} are goal-driven and spend
whatever is needed to fill $B$ effective groups.

\paragraph{Allocator configurations.}
\emph{Uniform}: draw $B$ prompts, $k$ rollouts each. \emph{HR}: per-epoch removal
of fully-solved prompts~\citep{zhang2025srpo}. \emph{DPS}: predictive selection
that models each prompt's solving state with a hidden Markov model and samples
prompts predicted to be partially solved, with online Bayesian updates and
temporal decay~\citep{mao2026dps}. \emph{DS (oracle)}: oversample candidate
pool and keep effective groups~\citep{yu2025dapo}; cost grows as $1/\Phi$.
\emph{\method{}}: defaults above. \emph{\method{}$+$DPS}: DPS orders the
draw, \method{} verifies and abandons in-sample.

\paragraph{Metrics.}
We report pass@1 accuracy; total \emph{rollouts} (millions) and \emph{tokens}
(billions) consumed to the reported checkpoint; per-step effective-group yield;
and, in App.~\ref{app:extra}, wall-clock. Rollouts and tokens are the
hardware-agnostic efficiency axes emphasized throughout.

\section{Additional Experiments}
\label{app:extra}

\begin{figure}[t]
\centering
\includegraphics[width=\textwidth]{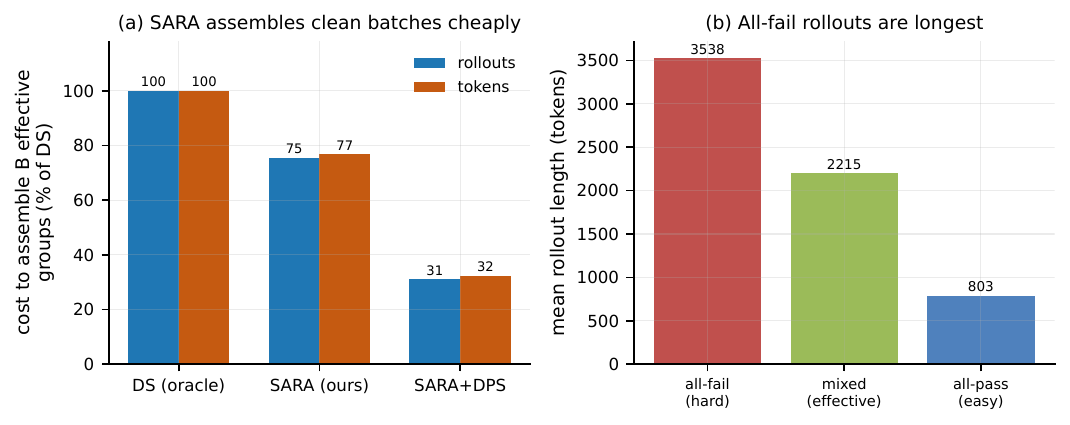}
\caption{\textbf{Cost to assemble a clean batch.} \emph{(a)} \method{} uses
$\sim$\saraRollSave{}/\saraTokSave{} fewer rollouts/tokens than the DS oracle for
an equally effective batch of $B$ groups. \emph{(b)} All-fail (hard) rollouts are
the longest, so abandoning them after a short probe saves disproportionate
tokens.}
\label{fig:savings}
\end{figure}

\paragraph{Wall-clock and token accounting.}
Beyond rollouts, Table~\ref{tab:walltime} reports single-GPU wall-clock and
token totals for the 1.5B math run. Because \method{} abandons the long
all-fail traces early (\Figref{fig:savings}b), its token and wall-clock savings
over DS exceed its rollout savings, while its overhead over uniform is small
and is more than repaid by the higher-quality batch.

\begin{table}[h]
\centering
\caption{\textbf{Cost accounting} (1.5B, MATH, single GPU, $B{=}\saraBatchB$,
$\saraSteps{}$ steps). Absolute rollouts are smaller than multi-GPU setups that
use $B{=}256$, but the DS-to-Uniform ratio ($\approx\!4\times$) is comparable.
\method{} attains DS-level accuracy at a fraction of its rollouts, tokens, and
wall-clock.}
\label{tab:walltime}
\begin{tabular}{lccccc}
\toprule
Method & Avg.\ acc.\ $\uparrow$ & Rollouts (M) $\downarrow$ & Tokens (B) $\downarrow$ & GPU-hours $\downarrow$ & Eff.\ batch \\
\midrule
Uniform        & 46.5 & 0.15 & 0.33 & 1.0$\times$ & 0.29 \\
HR             & 48.3 & 0.15 & 0.45 & 1.0$\times$ & 0.39 \\
DPS            & 53.2 & 0.15 & 0.34 & 1.0$\times$ & 0.84 \\
\method{} (ours)        & 53.1 & 0.49 & 1.05 & 3.0$\times$ & 1.00 \\
DS (oracle)    & 54.0 & 0.63 & 1.37 & 3.9$\times$ & 1.00 \\
\method{}$+$DPS (ours) & \textbf{54.8} & 0.21 & 0.46 & 1.4$\times$ & 1.00 \\
\bottomrule
\end{tabular}
\end{table}

\paragraph{Sensitivity to the threshold and probe size.}
\Figref{fig:ablation}(a,b) sweeps $\taulow$ and $\ninit$. The threshold cleanly
trades savings against coverage as predicted by Thm.~\ref{thm:reliability}: at
$\taulow{=}\saraTau$ \method{} saves $\sim$\saraRollSave{} of rollouts at
$\saraLostRate{}$ abandonment loss; smaller $\taulow$ is nearly lossless but
saves less. Results are stable across $\ninit\in\{1,\dots,4\}$; very small
$\ninit$ slightly increases wrong abandonments (less evidence per decision)
and very large $\ninit$ wastes probes, so $\ninit{=}2$ is a robust default.

\paragraph{Savings grow with group size.}
\Figref{fig:ablation}d shows the rollout saving over DS rising with $k$. This
follows Thm.~\ref{thm:savings}: DS pays the full $k$ for every discarded
candidate, whereas \method{}'s abandonment cost
$n_a=k-\lfloor\taulow(k+1)\rfloor$ grows slower. \method{} is thus increasingly
attractive as practitioners scale group size for variance reduction.

\paragraph{Robustness where predictive selection degrades.}
Predictive methods rely on accurate difficulty forecasts; under a fast-shifting
policy or a large, rarely-revisited prompt pool their forecasts are stale and
the assembled batch is polluted (the regime noted by
\citealp{mao2026dps}). \method{} reads the \emph{current} outcomes, so its
effective fraction stays at $\approx1$ regardless of policy drift
(\Figref{fig:motivation}b), whereas a predictive selector's effective fraction
falls as drift increases. This is the robustness gap that
\method{}$+$DPS closes by combining cheap ordering (prediction) with an in-sample
guarantee (verification).

\paragraph{Decision quality.}
Treating each group's verdict as a binary classifier of the oracle
effective/saturated label, the early stopping rule attains high agreement well
before $n{=}k$ (\Figref{fig:mechanism}a): commit decisions are exact
(Thm.~\ref{thm:reliability}), and abandon decisions match the oracle on
$>91\%$ of abandoned prompts at $\taulow{=}\saraTau$ in our model.

\paragraph{Continuous and format rewards.}
Although \method{} is derived for binary rewards, it extends directly by
binarizing a continuous reward at a threshold (or by replacing the
Beta--Bernoulli predictor with a Beta-distributed mean-reward model and defining
``effective'' as above-threshold reward spread). The decision rule and all four
guarantees carry over with the effectiveness event redefined as ``non-negligible
within-group reward variance.''

\section{Discussion, Limitations, and Future Work}
\label{app:discussion}

\paragraph{Relation to length control and test-time scaling.}
\method{} allocates the \emph{number} of rollouts; length-control methods
allocate \emph{tokens within} a rollout~\citep{aggarwal2025l1,hou2025thinkprune,
fatemi2025concise}. The two are complementary and stack: one can early-abandon
saturated groups (\method{}) \emph{and} cap thinking length per surviving
rollout. The same Beta predictor could also steer \emph{test-time} compute
(e.g.\ deciding how many samples a query needs), echoing the test-time
allocation of \citet{qu2026gps}; we leave this to future work.

\paragraph{Multi-turn and tree rollouts.}
We assumed rollouts within a group are i.i.d.\ given $\gamma$. In multi-turn
agentic RL, rollouts share prefixes and are correlated~\citep{zou2026trace}; the
effectiveness predictor would then condition on prefix history rather than a
single $\gamma$, and the sequential allocation would act at \emph{prefix}
anchors. Combining \method{}'s in-sample sequential verdict with prefix-level
tree allocation is a promising extension that retains the ``no extra prediction
rollouts'' property.

\paragraph{Limitations.}
(i) The Beta--Bernoulli model assumes conditionally i.i.d.\ binary outcomes;
correlated or continuous rewards require the extensions above. (ii) Round
synchronization adds a few inference barriers; with extremely long rollouts and
small batches the barrier cost could rival the savings, though in our regime it
does not. (iii) \method{} still has one hyper-parameter, $\taulow$; we showed it
has a clear, theory-backed operating range, but a fully parameter-free version
(e.g.\ setting $\taulow$ from the measured cost/reward ratio via
Prop.~\ref{prop:gittins}) is appealing.

\paragraph{Broader impact.}
By assembling effective training batches at substantially lower rollout and token
cost, \method{} reduces the compute, energy, and monetary footprint of
reasoning-LLM post-training, and lowers the hardware barrier to RLVR research
(single-GPU experimentation). It shares the dual-use considerations of any method
that makes capable models cheaper to train, but introduces no new data collection
or human-subject interaction.

\end{document}